%% file: main.tex
\documentclass[11pt]{article}

\usepackage{acl}

\usepackage{times}
\usepackage{latexsym}
\usepackage[T1]{fontenc}
\usepackage[utf8]{inputenc}
\usepackage{microtype}
\usepackage{inconsolata}
\usepackage{graphicx}
\usepackage{booktabs}
\usepackage{amsmath}
\usepackage{amssymb}
\usepackage{amsthm}
\usepackage{algorithm}
\usepackage{algpseudocode}
\usepackage{algorithmicx}
\usepackage{multirow}
\usepackage{enumitem}
\usepackage{subfigure}
\usepackage{wrapfig}
\usepackage[table]{xcolor}

\definecolor{PathRoute}{HTML}{6A0DAD}
\definecolor{GRPO}{HTML}{56C596}
\definecolor{OPCD}{HTML}{4DA5D9}
\definecolor{RLSD}{HTML}{E8793A}

\newcommand{\KL}{\mathrm{KL}}
\newcommand{\methodname}{\textsc{PathRouter}}

\title{PathRouter: Aligning Rewards with Retrieval Quality in Agentic Graph Retrieval-Augmented Generation}

\author{Bo Wang\textsuperscript{1}, Heyan Huang\textsuperscript{1}, Yaolin Li\textsuperscript{1}, Wei Tang\textsuperscript{2}, Yuan Zhang\textsuperscript{2}, Wenbo Li\textsuperscript{2}, \\ \textbf{Mingze Gao}\textsuperscript{1}, \textbf{Ge Shi}\textsuperscript{1},\textbf{Chong Feng}\textsuperscript{1}\\       
        \textsuperscript{1}Beijing Institute of Technology
        \\ \textsuperscript{2} Joy Future Academy
        }

\begin{document}
\maketitle

\input{sections/abstract}

\input{sections/introduction}

\input{sections/related_work}

\input{sections/method}

\input{sections/experiments}

\input{sections/conclusion}

\input{sections/limitations}

\bibliography{custom}

\appendix
\input{sections/appendix}

\clearpage
\section{Case Studies}
\label{sec:case_studies}

Tables~\ref{tab:case_mj}, \ref{tab:case_hotpotqa}, and \ref{tab:case_salary} present three representative case studies comparing Base, GRPO, and \methodname{} on multi-hop questions. Each case illustrates a distinct failure mode of outcome-only training: correct answer from parametric memory with no supporting evidence retrieved (Table~\ref{tab:case_mj}), correct answer via a hallucinated bridge entity (Table~\ref{tab:case_hotpotqa}), and exact answer recalled from parametric knowledge despite exhaustive but unsuccessful retrieval (Table~\ref{tab:case_salary}). In all three cases, \methodname{} reaches the answer through evidence-grounded reasoning chains with higher path overlap.

\input{case_study/case_004_mj_statue}
\input{case_study/case_hotpotqa_romantic_comedy}
\input{case_study/case_007_governor_salary}

\end{document}

%% file: sections/abstract.tex
\begin{abstract}
Agentic GraphRAG trains language-model agents to iteratively retrieve and reason over graph-structured evidence, enabling more accurate and context-aware decision-making by efficiently navigating complex information networks.
However, outcome-only reinforcement learning suffers from \textit{\textbf{answer-path reward aliasing}}, where correct answers may come from shortcuts rather than useful evidence paths. 
It also exhibits \textit{\textbf{search-update ambiguity}}, as scalar trajectory-level feedback does not indicate which retrieval actions to adjust.
To mitigate these shortcomings, we present \methodname{}, a path-aware training framework for agentic GraphRAG. 
\methodname{} jointly evaluates each trajectory along answer correctness and evidence-path overlap, yielding four trajectory categories with differentiated GRPO advantage scaling that suppresses shortcut reinforcement while preserving evidence-seeking behavior. 
For evidence-poor trajectories, a frozen gold-evidence teacher provides token-level KL guidance on reasoning and search-query tokens, excluding answer tokens to avoid direct response imitation. 
Experiments on six QA benchmarks across three model sizes show that \methodname{} consistently improves answer F1 and evidence-path overlap, achieving average F1 gains of 3.1 on 3B and 4.9 on 7B models compared to a strong baseline.
\end{abstract}

%% file: sections/introduction.tex
\section{Introduction}
\label{sec:introduction}

\begin{figure}[t]
\centering
\includegraphics[width=\linewidth]{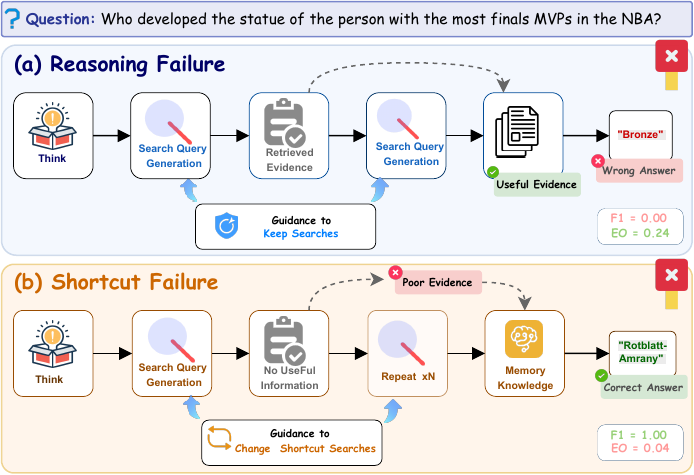}
\caption{Two failure modes.(a) reasoning failure with insufficient or misleading evidence, and (b) shortcut failure relying on parametric memory. Evidence-path overlap (EO), measuring token-level F1 between retrieved and gold evidence, highlights that modes receive the scalar reward but require different query-level updates.}
\label{fig:motivation}
\end{figure}

Retrieval-augmented generation (RAG) reduces hallucinations in large language models by grounding output in external knowledge~\citep{RAG,RAGsurvey,wang-etal-2024-qrmem}. 
Standard chunk-based pipelines, however, treat passages independently and fail to capture relational structure among entities.
Recent methods organize knowledge as entity-relation graphs, enabling multi-step retrieval over structured evidence paths~\citep{GraphRAG,LightRAG,PathRAG,HyperGraphRAG,HippoRAG2}, while agentic approaches model retrieval as iterative agent-environment interactions optimized via reinforcement learning~\citep{Search-R1,R1-Searcher,GraphR1}.

Despite these advances, existing agentic GraphRAG methods rely solely on answer correctness for trajectory rewards, ignoring whether retrieved evidence truly supports the answer. 
As illustrated in Figure~\ref{fig:motivation}, this gives rise to two failure modes: a correct answer may arise from parametric memory despite failed evidence retrieval, while an incorrect answer may occur even when key evidence is retrieved. This \textbf{answer-path reward aliasing} conflates trajectories of differing retrieval quality, and scalar trajectory-level feedback creates \textbf{search-update ambiguity}, leaving the agent without actionable guidance to improve its retrieval strategy.

Motivated by these failure modes, we propose \methodname{}, a path-aware training framework that jointly addresses answer-path reward aliasing and search-update ambiguity.
To resolve reward aliasing, \methodname{} evaluates each trajectory along two axes: answer correctness and evidence-path overlap (EO). 
By categorizing trajectories into four types based on these axes, the framework applies differentiated GRPO advantage scaling to suppress reinforcement of shortcut trajectories while preserving evidence-seeking behavior, ensuring that policy updates reflect retrieval quality rather than only final answer correctness.

To mitigate search-update ambiguity, \methodname{} selectively invokes a frozen gold-evidence teacher for evidence-poor trajectories. 
The teacher, prompted with gold supporting passages, provides token-level KL supervision on reasoning and search-query tokens, while excluding answer tokens to avoid direct response imitation. 
This guidance supplies actionable information for refining search strategies at the token level, allowing the agent to identify which entities, relations, or queries need adjustment to uncover missing evidence paths.

Through this design, \methodname{} enables agentic GraphRAG models to learn retrieval policies that are both \textbf{answer-accurate} and \textbf{evidence-faithful}. 
Experiments on six QA benchmarks across three model sizes demonstrate consistent improvements, with enhanced cross-dataset generalization and robust performance for both single-hop and multi-hop reasoning tasks. 
Our contributions are summarized as follows:
\begin{itemize}[nosep,leftmargin=1em]
    \item We identify answer-path reward aliasing as a failure mode for agentic GraphRAG and propose route-conditioned GRPO advantage scaling based on jointly evaluating answer correctness and evidence-path overlap.
    \item We introduce a selective gold-evidence teacher that applies token-level KL to reasoning and search-query tokens for evidence-poor trajectories, resolving search-update ambiguity without directly imitating final answers.
    \item We demonstrate consistent improvements over existing state-of-the-art methods across six QA benchmarks and three model sizes, with strong cross-dataset OOD transfer suggesting that path-aware training yields generalizable retrieval behavior rather than dataset-specific shortcuts.
\end{itemize}

%% file: sections/related_work.tex
\section{Related Work}
\label{sec:related_work}

\textbf{RAG and GraphRAG.}
Retrieval-augmented generation (RAG; \citealp{RAG}) grounds LLMs in external evidence, but chunk-based retrieval lacks explicit relational modeling~\citep{RAGsurvey}. GraphRAG methods address this by organizing knowledge as entity-relation graphs~\citep{GraphRAG,LightRAG,wang-etal-2023-boosting}, with path-based strategies exposing multi-hop reasoning chains via relational pruning~\citep{PathRAG} or iterative traversal~\citep{HopRAG}. However, these methods primarily optimize graph construction, path selection, or traversal heuristics; they do not directly train the generator to align its reasoning with annotated evidence paths. \methodname{} treats evidence-path overlap as a first-class training signal, routing updates based on whether the agent's retrieval covers the gold reasoning chain.

\textbf{Reinforcement Learning for Retrieval Agents.}
RL has become central to LLM reasoning, with GRPO~\citep{GRPO} providing scalable group-relative optimization popularized by DeepSeek-R1~\citep{DeepSeek-R1}. Search-R1~\citep{Search-R1} and R1-Searcher~\citep{R1-Searcher} apply GRPO to chunk-based retrieval-augmented reasoning, while Graph-R1~\citep{GraphR1} extends this to graph-structured knowledge. These systems primarily rely on outcome-level answer rewards, which can conflate retrieval quality with parametric knowledge and reinforce spurious shortcuts. \methodname{} introduces evidence-path overlap as a second diagnostic axis, enabling a 2$\times$2 trajectory taxonomy that distinguishes faithful from lucky success and applies route-conditioned advantage scaling accordingly.

\textbf{Knowledge Distillation and Sample Routing.}
On-policy distillation~\citep{DAgger} complements RL by providing token-level supervision on the student's own rollouts. Privileged-context teachers condition on gold evidence~\citep{OPCD,OPSD} or unify distillation with GRPO~\citep{KDRL,RLAD}. Sample routing directs trajectories to different signals: SRPO~\citep{SRPO} routes by correctness, GiGPO~\citep{GiGPO} decomposes advantages hierarchically, and RLSD~\citep{RLSD} highlights answer-token leakage risks. These methods generally apply teacher signals uniformly or route by answer correctness alone. \methodname{} conditions distillation on evidence-path overlap: a frozen teacher provides token-level KL only for evidence-poor trajectories, restricted to reasoning and query tokens, reducing answer leakage while targeting search-update ambiguity.

%% file: sections/method.tex
\begin{figure*}[t]
\centering
\includegraphics[width=\textwidth]{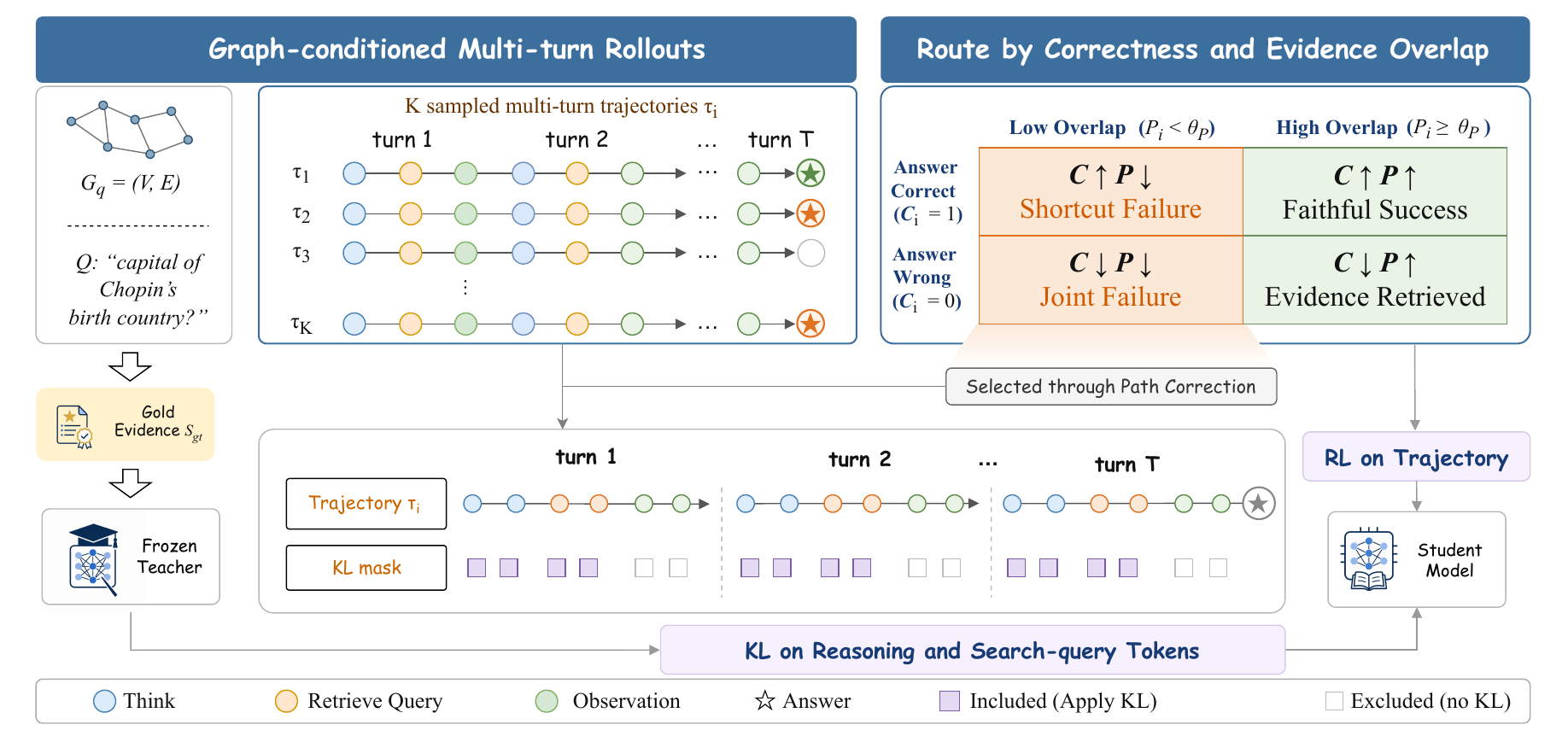}
\caption{Overview of \methodname{}. \textit{Top:} The student samples $K$ multi-turn trajectories per question on a graph-structured knowledge base $\mathcal{G}_q$, and each trajectory is routed into one of four categories by jointly evaluating answer correctness ($C_i$) and evidence-path overlap ($P_i$). \textit{Bottom:} Evidence-poor trajectories ($C\!\uparrow\!P\!\downarrow$, $C\!\downarrow\!P\!\downarrow$) are selected for path correction: a frozen teacher conditioned on gold evidence $S_{\text{gt}}$ provides token-level KL guidance on reasoning and query tokens ($\mathcal{T}_i^{\text{rq}}$), while knowledge and answer tokens are excluded. All trajectories receive route-weighted GRPO updates from the student model.}
\label{fig:overview}
\end{figure*}
\section{Approach}
\label{sec:method}
\vspace{-6pt}

\subsection{Task Formulation}
\label{sec:task_formulation}

We formalize agentic GraphRAG as a multi-turn agent-environment interaction following~\citet{GraphR1}. 
Given a knowledge graph $\mathcal{G}$ constructed from a document collection, an LLM-based agent $\pi_\theta$ interacts with a graph retrieval environment to answer a query $q$. 
At each turn $t$, the agent either emits a reasoning segment $r_t$ within \texttt{<think>} tags followed by a search query $q_t$ within \texttt{<query>} tags, or emits a final answer $a$ within \texttt{<answer>} tags to terminate the trajectory. The environment processes $q_t$ and returns retrieved passages within \texttt{<knowledge>} tags as the environment observation $o_t$. A reasoning trajectory is $\tau = ((r_1, q_1, o_1), \ldots, (r_T, q_T, o_T), a)$, where $T$ is the number of retrieval turns.

Building on this formulation, \methodname{} extends GRPO with a path-aware design. 
Trajectories are evaluated along answer correctness and evidence-path overlap (EO) to modulate GRPO advantage scaling, distinguishing shortcut answers from evidence-seeking ones. 
For evidence-poor trajectories, a selective gold-evidence teacher provides token-level KL guidance on reasoning and search-query steps without imitating final answers, encouraging retrieval that is both answer-accurate and evidence-faithful.

\subsection{Path-Aware Routing}
\label{sec:routing}

\methodname{} conditions the training signal on both answer correctness and evidence-path overlap.

\textbf{Trajectory Evaluation.}
For each trajectory $\tau_i$, we compute two diagnostic scores:
\begin{align}
C_i &= \mathbf{1}\bigl[\text{EM}(a_i, a^*) > 0 \;\lor\; \text{F1}(a_i, a^*) \geq \theta_C\bigr], \label{eq:ci} \\
P_i &= \frac{1}{|S_{\text{gt}}|}\sum_{s \in S_{\text{gt}}} \text{F1}\bigl(\text{retrieved}(\tau_i),\, s\bigr), \label{eq:pi}
\end{align}
where $C_i$ is binary answer correctness, and $P_i$ measures the average token-level F1 overlap between the concatenation of all passages retrieved along the trajectory and each gold supporting passage.
We determine correctness $C_i$ by either exact match or whether the token-level F1 exceeds a threshold $\theta_C$.
$P_i$ serves as a proxy for evidence-path quality: higher overlap indicates that the agent retrieved evidence relevant to the gold reasoning chain.

\textbf{Route Classification.}
Each trajectory is classified into one of four categories based on $C_i$ and an evidence-overlap threshold $\theta_P$:
\begin{equation}
\text{Route}(\tau_i)\!=\!\begin{cases}
C\!\uparrow P\!\uparrow & C_i = 1,\; P_i \geq \theta_P, \\
C\!\uparrow P\!\downarrow & C_i = 1,\; P_i < \theta_P, \\
C\!\downarrow P\!\uparrow & C_i = 0,\; P_i \geq \theta_P, \\
C\!\downarrow P\!\downarrow & C_i = 0,\; P_i < \theta_P.
\end{cases}
\end{equation}

\textbf{Route-Conditioned Advantage Scaling.}
Rather than assigning separate loss functions to each category, we modulate the GRPO advantage by a non-negative route-dependent weight:
\begin{equation}
\tilde{A}(\tau_i) = w^{\text{route}}(C_i, P_i) \cdot \hat{A}(\tau_i).
\end{equation}
The route weight modulates update magnitude, and the sign of the advantage remains determined by the group-normalized reward. The weights serve distinct purposes:
\begin{itemize}[nosep,leftmargin=1em]
    \item $C\!\uparrow P\!\uparrow$: \textit{Full weight} ($w_{\text{full}}$). The answer is correct and evidence overlap is high.
    \item $C\!\uparrow P\!\downarrow$: \textit{Reduced weight} ($w_{\text{down}}$). Answer is correct but evidence overlap is low, indicating a potential shortcut. 
    Reduced weight mitigates over-reinforcement of trajectories that achieve the correct answer without proper evidence retrieval.
    \item $C\!\downarrow P\!\uparrow$: \textit{Attenuated weight} ($w_{\text{preserve}}$). Evidence overlap is high but the answer is wrong, which is likely due to a final-reasoning failure. 
    Attenuated weight mitigates over-penalizing useful retrieval behavior.
    \item $C\!\downarrow P\!\downarrow$: \textit{Full weight} ($w_{\text{full}}$). Both answer correctness and evidence overlap are low.
\end{itemize}

\subsection{Distillation for Retrieval-Token}
\label{sec:teacher}

Route-conditioned scaling adjusts the magnitude of policy updates but leaves each trajectory with a single scalar advantage that does not specify which search actions should change. 
For evidence-poor trajectories ($P_i < \theta_P$) identified by the routing diagnostics in \S\ref{sec:routing}, we address this \textit{search-update ambiguity} by invoking a gold-evidence teacher that provides token-level distributional supervision on reasoning and search-query tokens, converting trajectory-level evidence failure into localized guidance for retrieval improvement. 
The teacher is used \textit{only during training}, whereas at inference the student acts without any privileged information.

Inspired by~\citet{OPCD}, we use the frozen reference model as the teacher. 
Crucially, the teacher does not generate its own trajectory during training, but instead scores the student's on-policy rollout under a privileged context. 
For each training sample $(q, \mathcal{G}, S_{\text{gt}}, a^*)$, the student first samples a trajectory $\tau_i$. 
At each token position $t$, both the student and teacher condition on the same response prefix $r_{<t}$ from this trajectory, differing only in their respective prompt context:
\begin{align}
\text{STUDENT:} &\quad \pi_\theta(\cdot \mid q, \mathcal{G}, r_{<t}), \\
\text{TEACHER:} &\quad \pi_{\text{ref}}(\cdot \mid q, \mathcal{G}, S_{\text{gt}}, r_{<t}),
\end{align}
where $S_{\text{gt}}$ is the set of gold supporting passages appended to the teacher's prompt. 
The teacher thus provides evidence-informed next-token guidance on the student's own on-policy trajectory, rather than sequence-level imitation of a separate teacher rollout. 
The frozen reference model ensures a stable distributional target throughout training.

In each response position $t$, both the student and the teacher observe the same prefix $r_{<t}$ but differ in prompt context. 
We retain the teacher's top-$K_{\text{vocab}}$ vocabulary tokens under $\pi_{\text{ref}}(\cdot \mid q, \mathcal{G}, S_{\text{gt}}, r_{<t})$, renormalized to a valid distribution $p_{\text{ref}}^{K}$. 
The student's logits are gathered at the same $K_{\text{vocab}}$ token indices and renormalized to $p_\theta^{K}$. 
This top-$K$ restriction reduces computational cost while focusing on the distillation of the teacher's high-confidence predictions.

\textbf{Retrieval-Token Masking.}
The KL supervision from the teacher is applied only to tokens within reasoning ($r_t$) and search-query ($q_t$) segments, i.e., the agent's own generated tokens, and excludes retrieved observation blocks ($o_t$) and the final answer ($a$). 
We denote the set of such positions in trajectory $\tau_i$ as $\mathcal{T}_i^{\text{rq}}$. 
This masking ensures that the teacher provides guidance on the agent's reasoning and evidence search, rather than directly instructing the final answer.

\textbf{Selective Teacher KL Loss.}
Trajectories with low evidence overlap ($P_i < \theta_P$) receive token-level KL supervision, while those with sufficient overlap are excluded from distillation to avoid interfering with effective evidence-seeking behavior. 
For eligible trajectories, we compute KL divergence over reasoning and query tokens:
\begin{equation}
\mathcal{L}_{\text{TKL}} = \frac{1}{|\mathcal{I}_{\text{low}}|}
\sum_{i \in \mathcal{I}_{\text{low}}} \frac{1}{|\mathcal{T}_i^{\text{rq}}|}
\sum_{t \in \mathcal{T}_i^{\text{rq}}} D_\KL\!\bigl(p_{\text{ref}}^{K} \,\big\|\, p_\theta^{K}\bigr)_{\!t},
\end{equation}
where $\mathcal{I}_{\text{low}} = \{i : P_i < \theta_P\}$ is the set of evidence-poor trajectories and the per-position KL is computed over the teacher's top-$K_{\text{vocab}}$ tokens under the teacher distribution. 
We use forward KL since the teacher distribution  provides a gold-evidence-conditioned target, and minimizing $D_\KL(p_{\text{ref}} \| p_\theta)$ encourages the student to cover the teacher's high-probability reasoning and search-query tokens.
In early training, most trajectories have low evidence overlap and would all qualify for teacher KL, which risks suppressing the on-policy exploration needed for the student to discover effective retrieval strategies. 
To mitigate this, we linearly warm up the KL coefficient $\lambda_T(s) = \bar{\lambda}_T \cdot \min(1, s / W)$ over the first $W$ steps, allowing the student to first learn from GRPO reward signals before the teacher constraint gradually takes effect.

\subsection{Training Objective}
\label{sec:objective}

\textbf{Reward Design.}
The reward combines task completion with evidence-path quality and exploration shaping:
\begin{equation}
R(\tau_i) = r_a + \alpha \, r_p + r_s,
\end{equation}
where $r_a = \text{F1}(a_i, a^*)$ measures answer quality, $r_p = P_i$ incorporates evidence-path overlap into the reward signal, and $r_s$ aggregates exploration shaping terms.
\begin{equation}
r_s = r_e + r_l + r_o + r_d.
\end{equation}
The exploration bonus $r_e$ rewards multi-turn retrieval for correctly-answered questions. The lazy penalty $r_l$ penalizes single-turn stops with insufficient evidence coverage. 
The timeout penalty $r_o$ penalizes reaching the maximum turn limit. 
The redundancy penalty $r_d$ discourages re-retrieving already-covered evidence.

\textbf{Routed GRPO Loss.}
For each training question, GRPO samples a group of $K$ trajectories from the behavior policy $\pi_{\theta_{\text{old}}}$. 
Each trajectory receives a reward $R(\tau_i)$, and its group-relative advantage $\hat{A}(\tau_i)$ is scaled by the weight of the route $\tilde{A}(\tau_i) = w^{\text{route}}(C_i, P_i) \cdot \hat{A}(\tau_i)$ from \S\ref{sec:routing}. 
The clipped policy-gradient loss is as follows:
\begin{equation}
\begin{split}
\mathcal{L}_{\text{GRPO}} = -\frac{1}{K}\sum_{i=1}^{K} \min\!\bigl(&\rho_i\,\tilde{A}(\tau_i),\\
&\text{clip}(\rho_i, 1\!-\!\epsilon, 1\!+\!\epsilon)\,\tilde{A}(\tau_i)\bigr),
\end{split}
\end{equation}
The overall training objective combines routed GRPO with selective teacher KL:
\begin{equation}
\mathcal{L} = \mathcal{L}_{\text{GRPO}} + \lambda_T(s) \cdot \mathcal{L}_{\text{TKL}}.
\end{equation}



%% file: sections/experiments.tex
\section{Experiments}
\label{sec:experiments}

We evaluate \methodname{} to answer the following research questions:
\textbf{RQ1}: Does \methodname{} improve answer accuracy and evidence quality over existing baselines? (\S\ref{sec:main_results})
\textbf{RQ2}: Which components drive path faithfulness, and how does routing change trajectory-level behavior? (\S\ref{sec:ablation_component}, \S\ref{sec:routing_analysis})
\textbf{RQ3}: How does teacher KL interact with model capacity, and do the learned retrieval strategies transfer across datasets? (\S\ref{sec:teacher_scale}, \S\ref{sec:ood})

\begin{table*}[t]
\centering
\caption{Main results across six QA benchmarks. \textbf{Bold}: best per model-size group; \underline{underline}: second best.
}
\label{tab:main}
\fontsize{7.2pt}{9pt}\selectfont
\setlength{\tabcolsep}{4.2pt}{
\begin{tabular}{l cc cc cc cc cc cc ccc}
\toprule
\multirow{2.5}{*}{\textbf{Method}}
  & \multicolumn{2}{c}{\textbf{2Wiki.}}
  & \multicolumn{2}{c}{\textbf{HotpotQA}}
  & \multicolumn{2}{c}{\textbf{MuSiQue}}
  & \multicolumn{2}{c}{\textbf{NQ}}
  & \multicolumn{2}{c}{\textbf{PopQA}}
  & \multicolumn{2}{c}{\textbf{TriviaQA}}
  & \multicolumn{3}{c}{\textbf{Avg.}} \\
\cmidrule(lr){2-3} \cmidrule(lr){4-5} \cmidrule(lr){6-7} \cmidrule(lr){8-9} \cmidrule(lr){10-11} \cmidrule(lr){12-13} \cmidrule(lr){14-16}
 & \textbf{F1} & \textbf{G-E}
 & \textbf{F1} & \textbf{G-E}
 & \textbf{F1} & \textbf{G-E}
 & \textbf{F1} & \textbf{G-E}
 & \textbf{F1} & \textbf{G-E}
 & \textbf{F1} & \textbf{G-E}
 & \textbf{EM} & \textbf{F1} & \textbf{G-E} \\
\midrule
\multicolumn{16}{c}{\cellcolor{gray!8}\textit{GPT-4o-mini$^\dagger$}} \\
NaiveGeneration          & 17.03 & 74.86 & 31.79 & 78.48 & 11.45 & 76.61 & 21.59 & 84.64 & 25.95 & \textbf{72.75} & 47.73 & 83.33 & 11.36 & 25.92 & 78.45 \\
StandardRAG              & \textbf{22.31} & 73.02 & \textbf{46.70} & \textbf{81.88} & 17.31 & 74.93 & \textbf{26.85} & 84.55 & \textbf{30.58} & 69.42 & \textbf{48.55} & 84.63 & \textbf{18.10} & \textbf{32.05} & 78.07 \\
GraphRAG                 & 16.02 & 72.81 & 31.67 & 77.37 & 15.14 & 74.43 & 20.31 & 82.36 & 20.92 & 65.88 & 45.13 & 82.76 & 12.50 & 24.87 & 75.94 \\
LightRAG                 & 16.59 & 71.94 & 30.70 & 73.42 & 14.39 & 73.75 & 19.09 & 80.20 & 20.47 & 67.76 & 40.18 & 81.60 & 9.77 & 23.57 & 74.78 \\
PathRAG                  & 12.42 & 67.19 & 23.12 & 71.81 & 11.49 & 69.94 & 20.01 & 81.99 & 15.65 & 60.58 & 37.44 & 80.94 & 7.03 & 20.02 & 72.08 \\
HippoRAG2                & 16.27 & 68.78 & 31.78 & 76.43 & 12.37 & 73.05 & 24.56 & \textbf{84.65} & 21.10 & 63.31 & 46.86 & 83.55 & 13.80 & 25.49 & 74.96 \\
HyperGraphRAG            & 21.14 & \textbf{76.76} & 37.46 & 80.50 & \textbf{20.40} & \textbf{79.29} & 22.95 & 81.22 & 29.48 & 70.55 & 44.95 & \textbf{85.20} & 13.15 & 29.40 & \textbf{78.92} \\
\midrule
\multicolumn{16}{c}{\cellcolor{gray!8}\textit{Qwen2.5-1.5B-Instruct$^\dagger$}} \\
NaiveGeneration          & 7.78 & 49.13 & 4.27 & 45.77 & 2.35 & 46.63 & 6.03 & 46.74 & 10.06 & 42.67 & 8.10 & 52.92 & 1.17 & 6.43 & 47.31 \\
StandardRAG              & 11.46 & 55.38 & 9.93 & 52.91 & 3.18 & 39.46 & 11.39 & 59.73 & 13.08 & 50.29 & 17.43 & 60.52 & 5.73 & 11.08 & 53.05 \\
SFT                      & 13.26 & 34.72 & 13.61 & 38.93 & 5.14 & 28.50 & 11.56 & 46.61 & 15.61 & 31.35 & 26.18 & 46.66 & 9.83 & 14.23 & 37.80 \\
R1                       & 26.28 & 47.48 & 20.07 & 44.43 & 4.84 & 39.12 & 16.75 & 45.95 & 21.36 & 44.50 & 34.78 & 48.59 & 14.19 & 20.68 & 45.01 \\
Search-R1                & 28.43 & 60.61 & 39.99 & 64.16 & 4.69 & 39.32 & 20.26 & 59.93 & 39.63 & 58.19 & 44.16 & 63.01 & 23.18 & 29.53 & 57.54 \\
R1-Searcher              & 28.01 & 58.81 & 41.50 & 61.54 & 6.26 & 38.31 & 36.86 & 60.79 & 38.37 & 56.02 & 42.57 & 61.24 & 23.70 & 32.26 & 56.12 \\
Graph-R1                 & 35.13 & 65.73 & 40.62 & 65.30 & 28.28 & 58.82 & 35.62 & 59.13 & 43.55 & 66.46 & \underline{57.36} & 70.83 & 31.90 & 40.09 & 64.38 \\
\rowcolor{blue!5}
\methodname{}            & \textbf{37.95} & \textbf{68.42} & \underline{46.38} & \underline{68.53} & \textbf{32.74} & \textbf{62.18} & \underline{38.02} & \textbf{63.70} & \underline{46.25} & \textbf{69.84} & \textbf{58.50} & \textbf{73.65} & \textbf{35.55} & \underline{43.31} & \textbf{67.72} \\
\rowcolor{blue!5}
\; \methodname{} w/o TKL & \underline{37.61} & \underline{67.50} & \textbf{50.17} & \textbf{69.81} & \textit{31.21} & \underline{60.94} & \textbf{39.57} & \underline{63.28} & \textbf{47.40} & \underline{68.22} & 56.05 & \underline{72.30} & \underline{34.51} & \textbf{43.67} & \underline{67.01} \\
\midrule
\multicolumn{16}{c}{\cellcolor{gray!8}\textit{Qwen2.5-3B-Instruct$^\dagger$}} \\
NaiveGeneration          & 7.59 & 55.00 & 11.16 & 53.75 & 3.67 & 54.00 & 8.90 & 57.18 & 10.89 & 49.08 & 10.89 & 48.16 & 3.26 & 8.85 & 52.86 \\
StandardRAG              & 12.52 & 60.01 & 15.41 & 62.51 & 2.92 & 50.40 & 10.69 & 65.13 & 14.70 & 57.25 & 21.92 & 68.43 & 3.39 & 13.03 & 60.62 \\
SFT                      & 12.40 & 52.31 & 16.48 & 51.35 & 5.04 & 51.31 & 11.23 & 58.20 & 16.95 & 46.42 & 33.02 & 59.98 & 9.64 & 15.85 & 53.26 \\
R1                       & 28.45 & 56.92 & 25.33 & 55.38 & 8.07 & 47.53 & 21.51 & 55.11 & 27.11 & 48.65 & 47.91 & 60.74 & 19.66 & 26.40 & 54.06 \\
Search-R1                & 38.04 & 54.39 & 43.84 & 69.32 & 7.65 & 46.43 & 37.96 & 52.90 & 38.67 & 63.74 & 47.99 & 60.37 & 28.65 & 35.69 & 57.86 \\
R1-Searcher              & 23.50 & 55.86 & 42.44 & 64.60 & 12.81 & 50.07 & 36.53 & 63.33 & 40.18 & 66.23 & 54.00 & 60.52 & 27.08 & 34.91 & 60.10 \\
Graph-R1                 & 57.56 & 76.45 & 56.75 & 77.46 & 40.51 & 67.84 & 44.75 & 69.92 & 45.65 & 71.27 & 62.31 & 75.01 & 42.45 & 51.26 & 72.99 \\
\rowcolor{blue!5}
\methodname{}            & \textbf{60.34} & \textbf{78.92} & \textit{62.20} & \textbf{79.58} & \textbf{45.02} & \textbf{70.43} & \textit{47.80} & \textbf{73.16} & \textit{48.04} & \textbf{74.30} & \textbf{62.50} & \textbf{77.82} & \textbf{44.48} & \textbf{54.32} & \textbf{75.70} \\
\rowcolor{blue!5}
\; \methodname{} w/o TKL & \underline{60.06} & \underline{78.14} & \textbf{62.75} & \underline{79.20} & \underline{44.51} & \underline{69.58} & \textbf{48.05} & \underline{72.84} & \textbf{48.15} & \underline{73.52} & \underline{61.81} & \underline{77.18} & \underline{44.02} & \underline{54.22} & \underline{75.08} \\
\midrule
\multicolumn{16}{c}{\cellcolor{gray!8}\textit{Qwen2.5-7B-Instruct$^\dagger$}} \\
NaiveGeneration          & 12.25 & 66.75 & 16.58 & 65.31 & 4.06 & 65.47 & 13.00 & 69.56 & 12.82 & 60.50 & 24.51 & 72.65 & 3.12 & 13.87 & 66.71 \\
StandardRAG              & 12.75 & 60.06 & 21.10 & 66.13 & 4.53 & 59.84 & 15.97 & 70.49 & 16.10 & 60.86 & 24.90 & 73.71 & 5.34 & 15.89 & 65.18 \\
SFT                      & 20.28 & 63.85 & 27.59 & 65.65 & 10.02 & 63.50 & 19.02 & 68.19 & 27.93 & 56.31 & 39.21 & 70.25 & 15.57 & 24.01 & 64.63 \\
R1                       & 30.99 & 59.19 & 37.05 & 60.12 & 14.53 & 49.39 & 28.45 & 57.63 & 30.35 & 53.38 & 57.33 & 66.73 & 25.91 & 33.12 & 57.74 \\
Search-R1                & 41.29 & 70.26 & 50.85 & 73.85 & 22.35 & 57.68 & 45.88 & 67.58 & 50.76 & 66.08 & 65.98 & 76.15 & 38.54 & 46.19 & 68.60 \\
R1-Searcher              & 33.96 & 69.61 & 46.36 & 74.56 & 16.63 & 59.05 & 44.93 & 68.54 & 47.12 & 66.74 & 64.76 & 75.95 & 34.51 & 42.29 & 69.08 \\
Graph-R1                 & 65.04 & 82.42 & 62.69 & 80.03 & 46.17 & 71.42 & 49.87 & 70.97 & 51.22 & 73.43 & \underline{71.93} & 79.11 & 48.57 & 57.82 & 76.23 \\
\rowcolor{blue!5}
\methodname{}            & \textbf{71.04} & \textbf{84.20} & \textbf{70.13} & \textbf{82.56} & \textbf{54.34} & \textbf{74.18} & \textbf{53.84} & \textbf{74.62} & \textbf{54.80} & \textbf{78.30} & \textbf{72.29} & \underline{81.48} & \textbf{53.26} & \textbf{62.74} & \textbf{78.72} \\
\rowcolor{blue!5}
\; \methodname{} w/o TKL & \underline{67.49} & \underline{83.54} & \underline{65.93} & \underline{81.40} & \underline{51.61} & \underline{73.26} & \underline{51.52} & \underline{73.84} & \underline{52.71} & \underline{76.48} & 71.89 & \textbf{82.32} & \underline{49.74} & \underline{60.36} & \underline{78.47} \\
\bottomrule
\end{tabular}}
\end{table*}

\subsection{Experimental Setup}
\textbf{Datasets.}
We follow the evaluation protocol of Graph-R1~\citep{GraphR1} and use six QA benchmarks:
three multi-hop datasets (\textbf{HotpotQA}~\citep{HotpotQA}, \textbf{2WikiMultiHopQA}~\citep{2WikiMultiHopQA}, and \textbf{MuSiQue}~\citep{Musique}) that provide gold supporting facts enabling path faithfulness evaluation, and three single-hop datasets (\textbf{NQ}~\citep{NQ}, \textbf{PopQA}~\citep{PopQA}, and \textbf{TriviaQA}~\citep{TriviaQA}) for broader coverage.

\textbf{Baselines.} Following Graph-R1~\citep{GraphR1}, we compare the following methods:

\textit{GPT-4o-mini-based methods as references:} \textbf{GraphRAG}~\citep{GraphRAG}, \textbf{LightRAG}~\citep{LightRAG}, \textbf{PathRAG}~\citep{PathRAG}, \textbf{HippoRAG2}~\citep{HippoRAG2}, and \textbf{HyperGraphRAG}~\citep{HyperGraphRAG}.

\textit{Non-retrieval approaches:} \textbf{NaiveGeneration} (direct LLM), \textbf{SFT}~\citep{SFT} (supervised fine-tuning), and \textbf{R1}~\citep{GRPO} (GRPO without retrieval).

\textit{Chunk-based retrieval:} \textbf{StandardRAG}~\citep{RAG}, \textbf{Search-R1}~\citep{Search-R1}, and \textbf{R1-Searcher}~\citep{R1-Searcher}.

\textit{Graph-based agentic retrieval:} \textbf{Graph-R1}~\citep{GraphR1} (our primary baseline).

\textbf{Evaluation Metrics.}
We report token-level \textbf{F1}, \textbf{Exact Match (EM)}, and \textbf{GPT-4o-mini evaluation (G-E)} for answer quality, along with \textbf{Supporting Fact F1 (SF-F1)} and \textbf{Unsupported Answer Rate (UAR, $\downarrow$)} for path faithfulness. Formal definitions are in Appendix~\ref{app:metrics}.

\subsection{Main Results}
\label{sec:main_results}

Table~\ref{tab:main} reports results across six datasets and three model sizes, revealing two consistent patterns.
\methodname{} consistently outperforms Graph-R1 across all model scales, improving 7B average F1 from 57.82 to 62.74, EM from 48.57 to 53.26, and G-E from 76.23 to 78.72.
The agreement across lexical, exact-match, and LLM-based metrics suggests the gains reflect more reliable answer production rather than surface-form effects.
Improvements are most pronounced on multi-hop benchmarks where faithful evidence composition is essential, with 7B F1 rising by 7.4 points on HotpotQA and 8.2 on MuSiQue.
The same pattern holds at smaller scales, with average F1 improving by 3.1 points at 3B and 3.2 at 1.5B, confirming that route-conditioned training mitigates answer-path reward aliasing regardless of model capacity.
Route conditioning remains scale-robust, while TKL is more capacity-sensitive. Section~\ref{sec:teacher_scale} discusses whether this reflects teacher mismatch or limited student capacity.

\begin{table*}[t]
\centering
\caption{Ablation Study on multi-hop datasets and NQ (Qwen2.5-7B-Instruct).}
\label{tab:ablation_component}
\fontsize{7pt}{8.5pt}\selectfont
\setlength{\tabcolsep}{2pt}{
\begin{tabular}{l cccc cccc cccc cccc}
\toprule
\multirow{2.5}{*}{\textbf{Variant}}
  & \multicolumn{4}{c}{\textbf{HotpotQA}}
  & \multicolumn{4}{c}{\textbf{2WikiMultiHopQA}}
  & \multicolumn{4}{c}{\textbf{MuSiQue}}
  & \multicolumn{4}{c}{\textbf{NQ}} \\
\cmidrule(lr){2-5} \cmidrule(lr){6-9} \cmidrule(lr){10-13} \cmidrule(lr){14-17}
 & \textbf{F1} & \textbf{SF-F1} & \textbf{UAR}$\downarrow$ & \textbf{Turns}
 & \textbf{F1} & \textbf{SF-F1} & \textbf{UAR}$\downarrow$ & \textbf{Turns}
 & \textbf{F1} & \textbf{SF-F1} & \textbf{UAR}$\downarrow$ & \textbf{Turns}
 & \textbf{F1} & \textbf{SF-F1} & \textbf{UAR}$\downarrow$ & \textbf{Turns} \\
\midrule
\rowcolor{blue!5}
\methodname{} (full)                  & \textbf{70.13} & \textbf{53.09} & \textbf{16.54} & 2.8 & \textbf{71.04} & \textbf{47.48} & \textbf{29.89} & 3.5 & \textbf{54.34} & \textbf{48.20} & \textbf{39.95} & 3.7 & \textbf{53.84} & \textbf{53.35} & \textbf{26.72} & 3.2 \\
\midrule
\multicolumn{17}{l}{\textit{Reward}} \\
\; w/o path reward ($r_p\!=\!0$)      & 69.47 & 47.23 & 20.51 & 2.7 & 70.28 & 41.86 & 34.17 & 3.4 & 53.58 & 42.53 & 44.06 & 3.6 & 53.12 & 48.04 & 30.48 & 3.1 \\
\; w/o exploration bonus ($r_e\!=\!0$) & 43.51 & 22.46 & 42.03 & 1.0 & 40.18 & 18.83 & 52.47 & 1.0 & 26.82 & 16.24 & 61.95 & 1.0 & 38.57 & 28.51 & 40.53 & 1.0 \\
\; w/o lazy penalty ($r_l\!=\!0$)      & 48.23 & 30.51 & 33.96 & 1.3 & 45.82 & 26.17 & 44.53 & 1.5 & 31.47 & 22.83 & 53.94 & 1.5 & 40.16 & 32.95 & 35.48 & 1.2 \\
\; w/o timeout penalty ($r_o\!=\!0$)   & 69.83 & 52.47 & 17.02 & 3.2 & 70.68 & 46.83 & 30.52 & 3.8 & 53.91 & 47.53 & 40.47 & 4.0 & 53.52 & 52.78 & 27.18 & 3.5 \\
\; w/o redundancy penalty ($r_d\!=\!0$) & 69.72 & 52.18 & 17.23 & 2.9 & 70.53 & 46.52 & 30.84 & 3.6 & 53.78 & 47.16 & 40.82 & 3.8 & 53.41 & 52.47 & 27.53 & 3.3 \\
\midrule
\multicolumn{17}{l}{\textit{Teacher KL}} \\
\; w/o teacher KL                     & 65.93 & 52.72 & 16.80 & 3.0 & 67.49 & 47.21 & 30.03 & 3.4 & 51.61 & 48.66 & 40.90 & 3.0 & 51.52 & 53.35 & 20.97 & 2.6 \\
\; w/o selective KL (uniform)         & 63.18 & 44.97 & 24.53 & 2.9 & 64.52 & 40.23 & 37.96 & 3.4 & 48.27 & 40.53 & 48.47 & 3.6 & 50.13 & 46.52 & 32.96 & 3.1 \\
\; w/o KL warmup (constant $\lambda$) & 66.83 & 50.47 & 20.53 & 4.5 & 67.91 & 44.18 & 35.04 & 4.8 & 50.22 & 44.53 & 45.82 & 5.0 & 51.38 & 49.96 & 31.47 & 4.2 \\
\midrule
\multicolumn{17}{l}{\textit{Routing}} \\
\; w/o route scaling ($w\!=\!1$)      & 66.82 & 49.17 & 19.53 & 2.8 & 67.93 & 43.48 & 33.82 & 3.4 & 51.23 & 43.96 & 44.02 & 3.6 & 51.62 & 49.47 & 29.96 & 3.1 \\
\; 1D routing: answer-only ($C_i$)    & 67.53 & 48.52 & 19.48 & 2.7 & 68.83 & 43.04 & 33.52 & 3.4 & 52.17 & 43.98 & 43.52 & 3.6 & 52.53 & 49.48 & 29.53 & 3.1 \\
\; 1D routing: path-only ($P_i$)      & 66.87 & 50.16 & 18.52 & 2.8 & 68.24 & 44.23 & 33.04 & 3.5 & 51.53 & 45.82 & 42.47 & 3.7 & 51.93 & 50.48 & 28.97 & 3.2 \\
\bottomrule
\end{tabular}}
\end{table*}

\subsection{Ablation Study}
\label{sec:ablation_component}

Table~\ref{tab:ablation_component} assesses how the reward design, teacher KL objective, and routing mechanism each contribute to path faithfulness.
In the reward group, removing the path reward $r_p$ primarily lowers SF-F1 while retaining much of the answer F1, indicating that this signal improves evidence quality rather than merely supporting answer prediction.
By contrast, removing either the exploration bonus $r_e$ or the lazy penalty $r_l$ substantially reduces retrieval turns to approximately one and causes broad degradation across all metrics, showing that both terms are essential for sustaining multi-step evidence seeking.
The timeout $r_o$ and redundancy $r_d$ penalties yield more modest but consistent regularization benefits.

In the teacher KL group, removing KL preserves multi-turn behavior but reduces answer F1, whereas applying KL uniformly underperforms selective KL, suggesting that teacher guidance is most effective when focused on evidence-poor trajectories.
In the routing group, removing route scaling or reducing the 2D taxonomy to a single axis degrades both F1 and SF-F1, demonstrating that answer correctness and evidence-path overlap provide complementary supervision signals.

\begin{figure*}[t]
\centering
\includegraphics[width=0.92\textwidth]{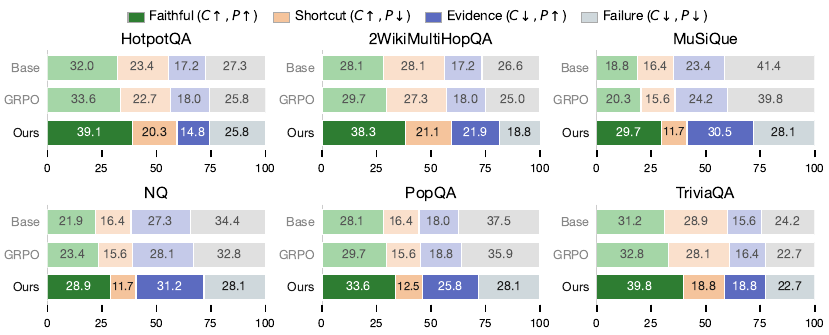}
\caption{Trajectory distribution across the four routing categories on all six datasets (Qwen2.5-7B-Instruct). Each stacked bar sums to 100\%. Saturated colors mark the \methodname{} (Ours) rows; lighter colors mark Base and GRPO. \methodname{} shifts mass from shortcut ($C\!\uparrow P\!\downarrow$) and failure ($C\!\downarrow P\!\downarrow$) toward faithful success ($C\!\uparrow P\!\uparrow$) and evidence-present reasoning failure ($C\!\downarrow P\!\uparrow$), reflecting improved retrieval quality across all answer outcomes.}
\label{fig:faithfulness}
\end{figure*}

\subsection{Routing and Trajectory Quality}
\label{sec:routing_analysis}

This analysis examines whether route-conditioned training promotes evidence-path exploration beyond improvements in answer F1.
Figure~\ref{fig:faithfulness} compares the converged 2$\times$2 trajectory distributions across all six datasets for Base, GRPO, and \methodname{}.
\methodname{} consistently reduces evidence-deficient trajectories, including shortcut success ($C\!\uparrow P\!\downarrow$) and joint failure ($C\!\downarrow P\!\downarrow$), while increasing faithful success ($C\!\uparrow P\!\uparrow$) on all datasets, with the largest gain on MuSiQue.
In addition, evidence-present reasoning failure ($C\!\downarrow P\!\uparrow$) increases on five of six datasets, indicating that trajectories previously failing at both retrieval and answering are converted into cases where relevant evidence is successfully retrieved, even when answer synthesis remains unresolved.
The overall shift from the $P\!\downarrow$ column to the $P\!\uparrow$ column demonstrates that \methodname{} improves evidence acquisition independently of answer correctness, supporting the conclusion that route-conditioned training encourages genuine evidence-path exploration rather than exploiting parametric answer priors.

\subsection{Teacher Scale Analyze}
\label{sec:teacher_scale}

\begin{table}[t]
\centering
\caption{Teachers across model sizes on MuSiQue.}
\label{tab:abl_teacher_cross}
\fontsize{7.5pt}{9.5pt}\selectfont
\setlength{\tabcolsep}{2.5pt}{
\begin{tabular}{lll cccc}
\toprule
\textbf{Student} & \textbf{Teacher} & \textbf{Type} & \textbf{F1} & \textbf{SF-F1} & \textbf{UAR}$\downarrow$ & \textbf{Turns} \\
\midrule
\multirow{3}{*}{1.5B}
 & 1.5B & Online           & 27.83 & 40.52 & 56.31 & 1.2 \\
 & \cellcolor{blue!5}1.5B & \cellcolor{blue!5}Frozen (ours) & \cellcolor{blue!5}32.74 & \cellcolor{blue!5}43.82 & \cellcolor{blue!5}43.15 & \cellcolor{blue!5}2.9 \\
 & 7B   & Frozen           & 15.83 & 26.41 & 63.27 & 4.8 \\
\midrule
\multirow{3}{*}{3B}
 & 3B   & Online           & 39.47 & 42.83 & 53.18 & 1.3 \\
 & \cellcolor{blue!5}3B & \cellcolor{blue!5}Frozen (ours) & \cellcolor{blue!5}45.02 & \cellcolor{blue!5}46.53 & \cellcolor{blue!5}40.82 & \cellcolor{blue!5}3.3 \\
 & 7B   & Frozen           & 33.61 & 36.82 & 54.17 & 4.5 \\
\midrule
\multirow{2}{*}{7B}
 & 7B   & Online           & 48.31 & 43.47 & 52.83 & 2.9 \\
 & \cellcolor{blue!5}7B & \cellcolor{blue!5}Frozen (ours) & \cellcolor{blue!5}\textbf{54.34} & \cellcolor{blue!5}\textbf{48.20} & \cellcolor{blue!5}\textbf{39.95} & \cellcolor{blue!5}3.7 \\
\bottomrule
\end{tabular}}
\end{table}

\begin{table}[t]
\centering
\caption{Schedule settings on MuSiQue.}
\label{tab:abl_teacher}
\fontsize{7.5pt}{9.5pt}\selectfont
\setlength{\tabcolsep}{2.5pt}{
\begin{tabular}{ll cccc}
\toprule
\textbf{Teacher} & \textbf{Schedule} & \textbf{F1} & \textbf{SF-F1} & \textbf{UAR}$\downarrow$ & \textbf{Turns} \\
\midrule
\multirow{3}{*}{Frozen $\pi_\text{ref}$}
 & Constant $\lambda$                & 50.22 & 44.53 & 45.82 & 5.0 \\
 & Full KL (no selective)            & 48.27 & 40.53 & 48.47 & 3.6 \\
 & \cellcolor{blue!5}Warmup $\lambda$ (ours) & \cellcolor{blue!5}\textbf{54.34} & \cellcolor{blue!5}\textbf{48.20} & \cellcolor{blue!5}\textbf{39.95} & \cellcolor{blue!5}3.7 \\
\midrule
\multirow{3}{*}{Online $\pi_\theta$}
 & Constant $\lambda$                & 44.83 & 38.47 & 55.23 & 2.8 \\
 & Full KL (no selective)            & 42.14 & 35.82 & 58.47 & 2.5 \\
 & Warmup $\lambda$                  & 48.31 & 43.47 & 52.83 & 2.9 \\
\bottomrule
\end{tabular}}
\end{table}

As shown in Table~\ref{tab:main}, the gains from route conditioning are broadly stable across model sizes, but teacher KL shows a stronger dependence on scale.
Comparing \methodname{} with its no-TKL variant reveals that teacher KL is not a generic regularizer but a capacity-dependent mechanism for resolving search-update ambiguity.
At 7B, teacher KL gains concentrated on multi-hop tasks where scalar rewards leave the most ambiguity about which search actions to reinforce.
At 3B the effect is modest but consistent; at 1.5B it becomes mixed, with slight degradation on NQ, suggesting that smaller students may be over-constrained by teacher distributions.
This pattern clarifies the division of labor: route-conditioned rewards address answer-path aliasing, while teacher KL reduces search-update ambiguity when the student has sufficient capacity to exploit it.

We examine whether the reduced benefit of teacher KL at smaller scales stems from limited teacher quality or limited student capacity, and how the KL signal should be scheduled.
To disentangle teacher quality from student capacity, Table~\ref{tab:abl_teacher_cross} compares online, same-size frozen, and cross-size frozen teachers across three model sizes on MuSiQue.
``Online'' uses the evolving student as its own teacher; ``Frozen'' uses a fixed reference checkpoint.
Online self-distillation is consistently inferior, with UAR above 51\% and reduced multi-turn exploration, confirming that an evolving teacher cannot provide a stable correction signal.
Same-size frozen teachers perform best at all scales, resolving search-update ambiguity while providing guidance that the student can effectively absorb.
Crucially, supervising the 1.5B student with a stronger 7B frozen teacher sharply degrades F1 and raises UAR above 63\%, indicating that the bottleneck is student capacity than teacher quality.

To identify the best scheduling strategy, Table~\ref{tab:abl_teacher} compares KL schedule variants in both frozen and online settings on the 7B model.
Warmup scheduling achieves the highest F1 while avoiding the over-exploration observed under constant $\lambda$, where the KL constraint dominates early training and inflates average turns.
Online variants remain worse than their frozen counterparts across all schedules.

\subsection{Cross-Dataset Transfer}
\label{sec:ood}

\begin{figure}[t]
\centering
\includegraphics[width=0.7\linewidth]{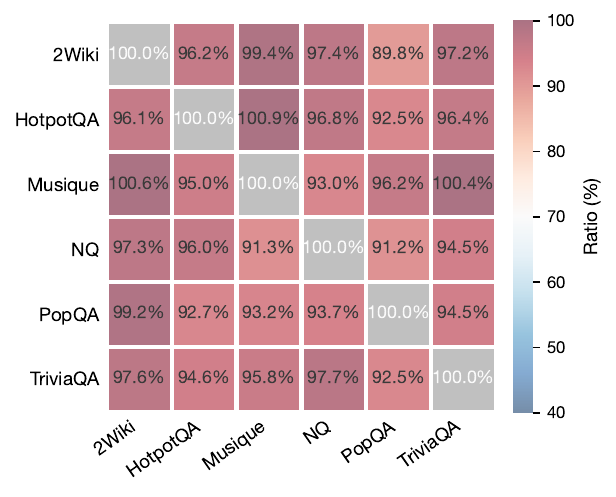}
\caption{Cross-dataset generalization of \methodname{} (Qwen2.5-7B-Instruct). Each cell shows the OOD ratio: $\text{F1}(\text{train}_i, \text{eval}_j) / \text{F1}(\text{train}_j, \text{eval}_j) \times 100\%$.}
\label{fig:ood}
\end{figure}

To assess whether path-faithful retrieval behavior generalizes beyond individual training distributions, we train \methodname{} on each dataset separately and evaluate on all others. Figure~\ref{fig:ood} reports the OOD generalization ratio for each train--eval pair. \methodname{} achieves a highly uniform transfer profile, with all off-diagonal ratios exceeding 89\% and an average ratio of 95.7\%, suggesting that path-aware routing and selective query guidance encourage retrieval strategies that are not tied to the distributional characteristics of any single training set. In comparison, Search-R1 and Graph-R1 achieve average ratios of 70.6\% and 85.8\% respectively (Appendix~\ref{app:ood}).

%% file: sections/conclusion.tex
\section{Conclusion}
\label{sec:conclusion}

We introduced \methodname{}, a path-aware training framework for agentic GraphRAG that addresses the credit assignment limitations of outcome-only reinforcement learning.
\methodname{} evaluates trajectories by answer correctness and evidence-path overlap, then modulates GRPO advantage weights to suppress shortcut reinforcement while preserving faithful-but-wrong explorations. 
For evidence-poor trajectories, a frozen gold-evidence teacher provides selective top-$K$ token-level KL guidance on reasoning and search-query tokens, with masking and warmup to prevent answer imitation and stabilize training. 
This path-conditioned design enables more targeted policy optimization than uniform distillation or outcome-only reinforcement.

%% file: sections/limitations.tex
\section*{Limitations}
\label{sec:limitations}

Our approach introduces additional hyperparameters for routing and teacher scheduling, which may require tuning when applied to smaller models or new domains. In addition, the path-aware reward encourages more exploration turns during training, and the teacher forward passes for selective KL add per-step cost. \methodname{} thus trades training efficiency for improved faithfulness and robustness.

\section*{Ethical Considerations}

The datasets and corpora used in our experiments are all open-access resources, and no private or sensitive information is involved. The models and APIs adopted in this work are obtained from publicly accessible platforms and are used in accordance with their respective terms and policies. Our study focuses on improving retrieval and reasoning over public knowledge corpora, and we do not anticipate any direct ethical concerns arising from this research.

%% file: sections/appendix.tex
\clearpage
\section{Evaluation Metrics}
\label{app:metrics}

We define the five evaluation metrics used throughout the paper.

\textbf{Token-level F1.}
Given a predicted answer $a$ and gold answer $a^*$, both tokenized into word sets, precision $P = |a \cap a^*| / |a|$ and recall $R = |a \cap a^*| / |a^*|$, then $\text{F1} = 2PR / (P + R)$. This is the standard SQuAD-style token-level F1~\citep{SQuAD}.

\textbf{Exact Match (EM).}
EM equals 1 if the normalized predicted answer string exactly matches the normalized gold answer, and 0 otherwise. Normalization includes lowercasing, removing articles, punctuation, and extra whitespace.

\textbf{Supporting Fact F1 (SF-F1).}
SF-F1 measures the average token-level F1 between the trajectory's retrieved passages and the gold supporting facts $S_{\text{gt}}$. Formally, $\text{SF-F1} = P_i = \frac{1}{|S_{\text{gt}}|}\sum_{s \in S_{\text{gt}}} \text{F1}(\text{retrieved}(\tau_i), s)$. This metric is equivalent to the evidence-overlap score $P_i$ defined in \S\ref{sec:routing}.

\textbf{Unsupported Answer Rate (UAR, $\downarrow$).}
UAR measures the fraction of answer tokens that are not grounded in the retrieved evidence. For each answer token $w \in a_i$, we check whether $w$ appears in any retrieved passage; $\text{UAR} = |\\{w \in a_i : w \notin \text{retrieved}(\tau_i)\\}| / |a_i|$. Lower UAR indicates better evidence grounding.

\textbf{GPT-4o-mini Evaluation (G-E).}
Following~\citet{GraphR1}, we use GPT-4o-mini as an LLM judge to score answer quality on a 0--100 scale. The judge receives the question, gold answer, and predicted answer, and evaluates correctness, completeness, and relevance. We report the average score across all test questions.

\section{Implementation Details}
\label{app:implementation}

We train with the veRL framework with group size $K=32$ and maximum $T=5$ retrieval turns. At inference, we increase the maximum turn limit to $T_{\text{inf}}=10$ to allow the agent to perform more thorough retrieval when needed; the timeout penalty $r_o$ is only applied during training.
Table~\ref{tab:hyperparams} lists the complete hyperparameter configuration.

\begin{table}[h]
\centering
\caption{Hyperparameter configuration for \methodname{}.}
\label{tab:hyperparams}
\small
\setlength{\tabcolsep}{3pt}
\begin{tabular}{llc}
\toprule
\textbf{Category} & \textbf{Hyperparameter} & \textbf{Value} \\
\midrule
\multirow{6}{*}{Training}
 & Learning rate & $1 \times 10^{-6}$ \\
 & Batch size & 128 \\
 & Group size $K$ & 32 \\
 & Max retrieval turns $T$ & 5 \\
 & Clip range $\epsilon$ & 0.2 \\
 & Training epochs & 3 \\
\midrule
\multirow{4}{*}{Teacher}
 & Top-$K$ vocab $K_{\text{vocab}}$ & 128 \\
 & KL coefficient $\bar{\lambda}_T$ & 0.1 \\
 & KL warmup steps $W$ & 20 \\
 & Ref.\ KL coefficient $\beta$ & 0.01 \\
\midrule
\multirow{4}{*}{Routing}
 & Answer threshold $\theta_C$ & 0.5 (F1) \\
 & Evidence threshold $\theta_P$ & 0.1 \\
 & $w_{\text{full}}$ / $w_{\text{down}}$ / $w_{\text{preserve}}$ & 1.0 / 0.5 / 0.3 \\
\midrule
\multirow{6}{*}{Reward}
 & Path weight $\alpha$ & 0.5 \\
 & Exploration bonus $r_e$ & 0.2 \\
 & Lazy penalty $r_l$ & $-0.25$ \\
 & Timeout penalty $r_o$ & $-0.5$ \\
 & Redundancy penalty $r_d$ & $-0.1$ \\
 & Base reward $R_0$ & $-1$ \\
\bottomrule
\end{tabular}
\end{table}

\section{Dataset Details}
\label{app:datasets}

\begin{table}[h]
\centering
\footnotesize
\begin{tabular}{lccc}
\toprule
\textbf{Dataset} & \textbf{Train} & \textbf{Dev} & \textbf{Test} \\
\midrule
HotpotQA & 90,564 & 7,405 & 500 \\
2WikiMultiHopQA & 167,454 & 12,576 & 500 \\
MuSiQue & 19,938 & 2,417 & 500 \\
\bottomrule
\end{tabular}
\caption{Dataset statistics.}
\label{tab:datasets}
\end{table}

\textbf{HotpotQA}~\citep{HotpotQA} provides sentence-level supporting facts for each question. We use these passages as gold evidence $S_{\text{gt}}$ for teacher conditioning and evidence-overlap computation. Each question is associated with 10 Wikipedia paragraphs (2 gold + 8 distractors) and requires 2-hop reasoning.

\textbf{2WikiMultiHopQA}~\citep{2WikiMultiHopQA} provides evidence information containing reasoning paths. We extract supporting passages from the evidence annotations as $S_{\text{gt}}$.

\textbf{MuSiQue}~\citep{Musique} provides decomposed sub-questions and their answers. We identify the paragraphs answering each sub-question as $S_{\text{gt}}$.

\section{Additional Experimental Results}
\label{app:additional}

\subsection{KL Selection Strategy and Threshold Sensitivity}
\label{app:kl_selection}

Table~\ref{tab:abl_selection} validates the KL selection design along two dimensions: trajectory selection and token masking. For trajectory selection, low-$P_i$ filtering outperforms content-based and uniform sampling, confirming that evidence overlap is the right criterion for identifying trajectories that benefit from teacher guidance. For token masking, restricting to query tokens alone slightly degrades SF-F1, while extending KL to answer tokens increases UAR from 39.95 to 46.23, showing that answer-token imitation creates shortcuts.

\begin{table}[h]
\centering
\caption{KL selection strategy on MuSiQue (Qwen2.5-7B-Instruct). All use frozen teacher with warmup $\lambda$.}
\label{tab:abl_selection}
\fontsize{7.5pt}{9.5pt}\selectfont
\setlength{\tabcolsep}{2.5pt}{
\begin{tabular}{l cccc}
\toprule
\textbf{Selection Strategy} & \textbf{F1} & \textbf{SF-F1} & \textbf{UAR}$\downarrow$ & \textbf{Turns} \\
\midrule
\cellcolor{blue!5}Low $P_i$ only (ours) & \cellcolor{blue!5}\textbf{54.34} & \cellcolor{blue!5}\textbf{48.20} & \cellcolor{blue!5}\textbf{39.95} & \cellcolor{blue!5}3.7 \\
Content-based sampling              & 51.83 & 44.96 & 43.52 & 3.5 \\
Uniform sampling in $P_i$ space     & 50.62 & 43.81 & 44.97 & 3.4 \\
Low $P_i$, query tokens only        & 53.17 & 46.53 & 41.82 & 3.6 \\
Low $P_i$, query + answer tokens    & 52.48 & 43.96 & 46.23 & 3.5 \\
\bottomrule
\end{tabular}}
\end{table}

Table~\ref{tab:threshold_sensitivity_main} reports threshold sensitivity. Performance is stable in $[0.05, 0.2]$ but degrades at $\theta_P\!=\!0.3$, where 88\% of trajectories receive teacher KL, approaching the uniform setting.

\begin{table}[h]
\centering
\caption{Sensitivity of $\theta_P$ on MuSiQue (Qwen2.5-7B-Instruct), with $\theta_C\!=\!0.5$ fixed.}
\label{tab:threshold_sensitivity_main}
\fontsize{7.5pt}{9.5pt}\selectfont
\setlength{\tabcolsep}{4pt}{
\begin{tabular}{c cccc}
\toprule
$\theta_P$ & \textbf{F1} & \textbf{SF-F1} & \textbf{UAR}$\downarrow$ & \textbf{KL-elig.\%} \\
\midrule
0.05 & 53.48 & 47.53 & 40.82 & 42 \\
\cellcolor{blue!5}0.1 (ours) & \cellcolor{blue!5}\textbf{54.34} & \cellcolor{blue!5}\textbf{48.20} & \cellcolor{blue!5}\textbf{39.95} & \cellcolor{blue!5}58 \\
0.2  & 53.81 & 47.76 & 40.47 & 75 \\
0.3  & 51.93 & 45.52 & 42.98 & 88 \\
\bottomrule
\end{tabular}}
\end{table}

\subsection{How does route quality evolve during training?}

\begin{figure}[t]
\centering
\subfigure[Route category proportions]{
    \includegraphics[width=0.92\columnwidth]{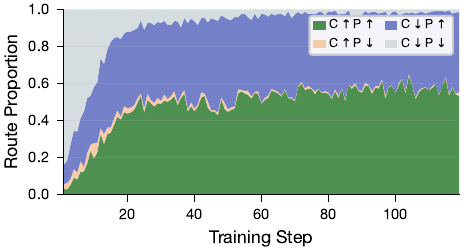}
    \label{fig:dynamics_a}
}
\subfigure[Training metrics]{
    \includegraphics[width=0.92\columnwidth]{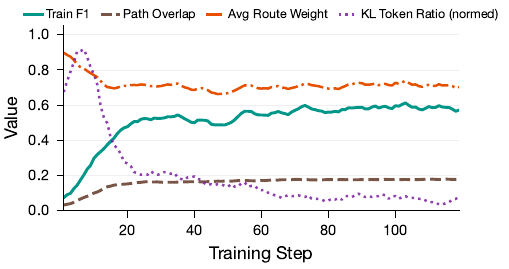}
    \label{fig:dynamics_b}
}
\caption{Training dynamics on HotpotQA (Qwen2.5-3B, 119 steps). (a)~Proportion of the four route categories over training: faithful trajectories ($C\!\uparrow\!P\!\uparrow$) grow steadily while joint failures ($C\!\downarrow\!P\!\downarrow$) vanish within the first 20 steps. (b)~Train F1 and Path Overlap rise together; Avg Route Weight stabilizes as the route distribution converges; KL Token Ratio decays as fewer trajectories remain evidence-poor.}
\label{fig:dynamics}
\end{figure}

Figure~\ref{fig:dynamics} tracks route proportions and training metrics over 119 steps on HotpotQA (Qwen2.5-3B).
Faithful trajectories ($C\!\uparrow P\!\uparrow$) grow steadily while joint failures vanish within the first 20 steps.
The KL token ratio decays naturally as fewer trajectories remain evidence-poor, creating an implicit curriculum from teacher-guided query correction to on-policy reinforcement.

\subsection{Baseline Cross-Dataset Transfer}
\label{app:ood}

\begin{figure}[t]
\centering
\subfigure[Search-R1]{
    \includegraphics[width=0.45\columnwidth]{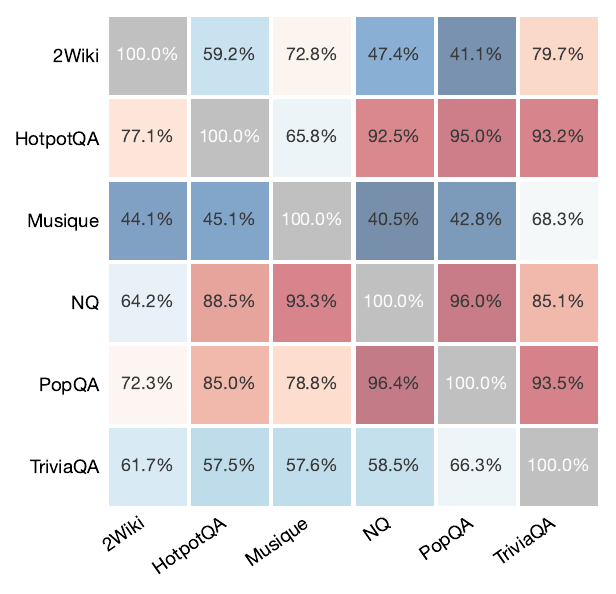}
    \label{fig:ood_search_r1}
}
\hfill
\subfigure[Graph-R1]{
    \includegraphics[width=0.45\columnwidth]{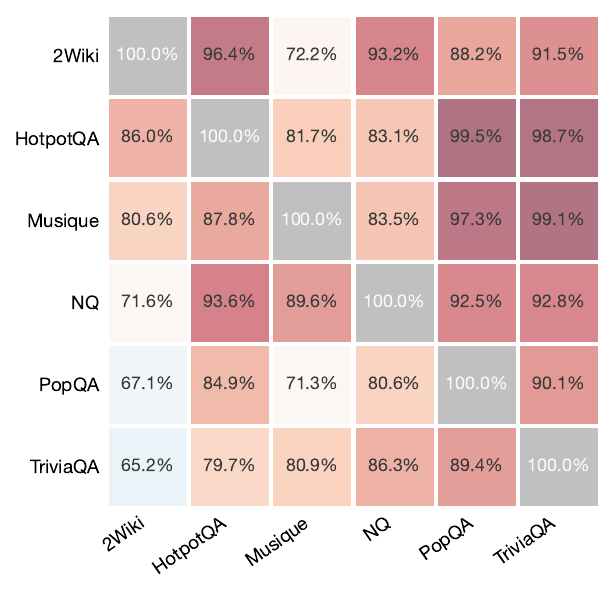}
    \label{fig:ood_graph_r1}
}
\caption{Cross-dataset generalization for baselines. Search-R1 uses 1.5B as reported by~\citet{GraphR1}; Graph-R1 uses Qwen2.5-7B-Instruct. Each cell shows the OOD ratio: $\text{F1}(\text{train}_i, \text{eval}_j) / \text{F1}(\text{train}_j, \text{eval}_j) \times 100\%$. Diagonal (gray) = in-domain.}
\label{fig:ood_baselines}
\end{figure}

Figure~\ref{fig:ood_baselines} shows the cross-dataset OOD generalization heatmaps for Search-R1 and Graph-R1. Search-R1 exhibits substantial OOD degradation with an average ratio of 70.6\%, indicating that outcome-only training tends to overfit to dataset-specific answer patterns. Graph-R1 improves transfer to 85.8\% through graph-structured retrieval, yet still shows notable drops on several cross-dataset pairs. In contrast, \methodname{} achieves an average ratio of 95.7\% (Figure~\ref{fig:ood}).

%% file: case_study/case_004_mj_statue.tex

\begin{table*}[t]
\centering
\caption{Case study on a 2-hop question from MuSiQue. \textbf{Question:} ``Who developed the statue of the person with the most finals MVPs in the NBA?'' \textbf{Ground truth:} Julie Rotblatt-Amrany. Base misidentifies the bridge entity (LeBron instead of MJ) and answers with the material rather than the sculptor. GRPO reaches the correct answer from parametric memory without retrieving supporting evidence (PO\,=\,0.109), while Ours retrieves the evidence in 3 turns (PO\,=\,0.346).}
\label{tab:case_mj}
\fontsize{7.5pt}{9pt}\selectfont
\setlength{\tabcolsep}{3pt}
\begin{tabular}{p{1.2cm} p{0.4cm} p{12cm}}
\toprule
\textbf{Method} & \textbf{Turn} & \textbf{Content} \\
\midrule
Base & 1
  & \textcolor{teal}{\texttt{<think>}} First, I need to identify the person with the most Finals MVP awards in NBA history. \textcolor{teal}{\texttt{</think>}} \\
& & \textcolor{blue}{\texttt{<query>}} Who has the most NBA Finals MVP awards? \textcolor{blue}{\texttt{</query>}} \\
& & \textcolor{gray}{\texttt{<knowledge>}} LeBron's honors include 3 Finals MVPs (2012, 2013, 2016)\ldots\ Willis Reed, Abdul-Jabbar, Bird, Olajuwon, Kobe each won twice\ldots \textcolor{gray}{\texttt{</knowledge>}} \\[3pt]
& 2
  & \textcolor{teal}{\texttt{<think>}} Based on the information, LeBron James has the most Finals MVP awards with 3. \textcolor{teal}{\texttt{</think>}} \\
& & \textcolor{blue}{\texttt{<query>}} Who developed the statue of LeBron James? \textcolor{blue}{\texttt{</query>}} \\
& & \textcolor{gray}{\texttt{<knowledge>}} Career highlights of LeBron James listed; no statue information found. \textcolor{gray}{\texttt{</knowledge>}} \\[3pt]
& 3
  & \textcolor{teal}{\texttt{<think>}} No statue for LeBron found. The Michael Jordan statue might be relevant. \textcolor{teal}{\texttt{</think>}} \\
& & \textcolor{blue}{\texttt{<query>}} What is the statue of Michael Jordan made of? \textcolor{blue}{\texttt{</query>}} \\
& & \textcolor{gray}{\texttt{<knowledge>}} The MJ statue ``The Spirit'' is a bronze by Omri Amrany and Julie Rotblatt-Amrany at the United Center. \textcolor{gray}{\texttt{</knowledge>}} \\[3pt]
& 4
  & \textcolor{teal}{\texttt{<think>}} The Michael Jordan statue is made of bronze. \textcolor{teal}{\texttt{</think>}} \\
& & \textcolor{red!70!black}{\texttt{<answer>}} Bronze \textcolor{red!70!black}{\texttt{</answer>}} \\[2pt]
& & \colorbox{gray!10}{\scriptsize \textbf{F1\,=\,0.000} ~|~ \textbf{PO\,=\,0.265} ~|~ \textbf{Turns\,=\,4} ~|~ Wrong sub-question (LeBron $\to$ material instead of sculptor)} \\
\midrule
GRPO & 1
  & \textcolor{teal}{\texttt{<think>}} I need to first find out who has the most Finals MVP awards in the NBA. \textcolor{teal}{\texttt{</think>}} \\
& & \textcolor{blue}{\texttt{<query>}} Who has the most NBA Finals MVP awards? \textcolor{blue}{\texttt{</query>}} \\
& & \textcolor{gray}{\texttt{<knowledge>}} LeBron: 3 Finals MVPs; Reed, Abdul-Jabbar, Bird, Olajuwon, Kobe each won twice. \textcolor{gray}{\texttt{</knowledge>}} \\[3pt]
& 2
  & \textcolor{teal}{\texttt{<think>}} LeBron James has the most Finals MVP awards with 3. \textcolor{teal}{\texttt{</think>}} \\
& & \textcolor{blue}{\texttt{<query>}} Who developed the statue of LeBron James? \textcolor{blue}{\texttt{</query>}} \\
& & \textcolor{gray}{\texttt{<knowledge>}} LeBron career highlights only; no statue information. \textcolor{gray}{\texttt{</knowledge>}} \\[3pt]
& \multicolumn{2}{l}{\quad\textcolor{gray!60}{\dotfill~\textit{Turns 3--7: five repeated queries about ``LeBron James statue'' with no relevant results}~\dotfill}} \\[3pt]
& 8
  & \textcolor{teal}{\texttt{<think>}} \textit{(forced)} The statue at Rocket Mortgage FieldHouse was created by Omri Amrany and Julie Rotblatt-Amrany, who also created the Michael Jordan statue. \textcolor{teal}{\texttt{</think>}} \\
& & \textcolor{red!70!black}{\texttt{<answer>}} Omri Amrany and Julie Rotblatt-Amrany \textcolor{red!70!black}{\texttt{</answer>}} \\[2pt]
& & \colorbox{gray!10}{\scriptsize \textbf{F1\,=\,0.571} ~|~ \textbf{PO\,=\,0.109} ~|~ \textbf{Turns\,=\,8} ~|~ Correct answer but from parametric memory; \textbf{no evidence retrieved}} \\
\midrule
Ours & 1
  & \textcolor{teal}{\texttt{<think>}} First, I need to find out who has the most Finals MVP awards in the NBA. \textcolor{teal}{\texttt{</think>}} \\
& & \textcolor{blue}{\texttt{<query>}} Who has the most NBA Finals MVP awards? \textcolor{blue}{\texttt{</query>}} \\
& & \textcolor{gray}{\texttt{<knowledge>}} LeBron: 3 Finals MVPs; Reed, Abdul-Jabbar, Bird, Olajuwon, Kobe each won twice\ldots\ Michael Jordan: 6 Finals MVPs (record). \textcolor{gray}{\texttt{</knowledge>}} \\[3pt]
& 2
  & \textcolor{teal}{\texttt{<think>}} Michael Jordan has the most Finals MVP awards with six. \textcolor{teal}{\texttt{</think>}} \\
& & \textcolor{blue}{\texttt{<query>}} Who developed the statue of Michael Jordan? \textcolor{blue}{\texttt{</query>}} \\
& & \textcolor{gray}{\texttt{<knowledge>}} The MJ statue ``The Spirit'' is a bronze by \textbf{Omri Amrany and Julie Rotblatt-Amrany} at the United Center. \textcolor{gray}{\texttt{</knowledge>}} \\[3pt]
& 3
  & \textcolor{teal}{\texttt{<think>}} The Michael Jordan statue was created by Omri Amrany and Julie Rotblatt-Amrany. \textcolor{teal}{\texttt{</think>}} \\
& & \textcolor{red!70!black}{\texttt{<answer>}} Omri Amrany and Julie Rotblatt-Amrany \textcolor{red!70!black}{\texttt{</answer>}} \\[2pt]
& & \colorbox{gray!10}{\scriptsize \textbf{F1\,=\,0.571} ~|~ \textbf{PO\,=\,0.346} ~|~ \textbf{Turns\,=\,3} ~|~ Correct reasoning chain; \textbf{answer grounded in retrieved evidence}} \\
\bottomrule
\end{tabular}
\end{table*}

%% file: case_study/case_hotpotqa_romantic_comedy.tex

\begin{table*}[t]
\centering
\caption{Case study on a 2-hop question from HotpotQA. \textbf{Question:} ``Who directed a romantic comedy starring a `7th Heaven' star and Ludacris?'' \textbf{Ground truth:} Garry Marshall. GRPO reaches the correct answer but relies on a hallucinated bridge entity (Halle Berry was never on 7th Heaven), whereas Ours identifies the valid bridge (Barret Swatek) through evidence-grounded retrieval.}
\label{tab:case_hotpotqa}
\fontsize{7.5pt}{9pt}\selectfont
\setlength{\tabcolsep}{3pt}
\begin{tabular}{p{1.2cm} p{0.4cm} p{12cm}}
\toprule
\textbf{Method} & \textbf{Turn} & \textbf{Content} \\
\midrule
Base & 1
  & \textcolor{teal}{\texttt{<think>}} I need to identify the ``7th Heaven'' star who starred in a romantic comedy alongside Ludacris, and then find the director. \textcolor{teal}{\texttt{</think>}} \\
& & \textcolor{blue}{\texttt{<query>}} Who starred in a romantic comedy with Ludacris, and is from the show ``7th Heaven''? \textcolor{blue}{\texttt{</query>}} \\
& & \textcolor{gray}{\texttt{<knowledge>}} ``How Low is a song by American rapper Ludacris.'' / ``She is best known for roles as Sandy Jameson in 7th Heaven, Amy Sanders on Lizzie McGuire\ldots'' \textcolor{gray}{\texttt{</knowledge>}} \\[3pt]
& 2
  & \textcolor{teal}{\texttt{<think>}} From the information provided, Mackenzie Rosman is a ``7th Heaven'' star. \textcolor{teal}{\texttt{</think>}} \\
& & \textcolor{blue}{\texttt{<query>}} Who directed the romantic comedy starring Mackenzie Rosman and Ludacris? \textcolor{blue}{\texttt{</query>}} \\
& & \textcolor{gray}{\texttt{<knowledge>}} ``New Year's Eve stars Halle Berry, Jessica Biel, Jon Bon Jovi, Abigail Breslin, Ludacris\ldots'' (no Mackenzie Rosman connection found) \textcolor{gray}{\texttt{</knowledge>}} \\[3pt]
& \multicolumn{2}{l}{\quad\textcolor{gray!60}{\dotfill~\textit{Turns 3--7: repeated ``Mackenzie Rosman and Ludacris'' queries; wrong bridge entity}~\dotfill}} \\[3pt]
& 8
  & \textcolor{teal}{\texttt{<think>}} Given the lack of relevant information, I cannot determine the director. \textcolor{teal}{\texttt{</think>}} \\
& & \textcolor{red!70!black}{\texttt{<answer>}} No information available. \textcolor{red!70!black}{\texttt{</answer>}} \\[2pt]
& & \colorbox{gray!10}{\scriptsize \textbf{F1\,=\,0.000} ~|~ \textbf{PO\,=\,0.037} ~|~ \textbf{Turns\,=\,8} ~|~ Fixated on wrong bridge entity; stuck in retrieval loop} \\
\midrule
GRPO & 1
  & \textcolor{teal}{\texttt{<think>}} I need to find out which romantic comedy starred a ``7th Heaven'' actor and Ludacris, and then identify the director. \textcolor{teal}{\texttt{</think>}} \\
& & \textcolor{blue}{\texttt{<query>}} Who starred in a romantic comedy with Ludacris? \textcolor{blue}{\texttt{</query>}} \\
& & \textcolor{gray}{\texttt{<knowledge>}} ``New Year's Eve stars Halle Berry, Jessica Biel, Jon Bon Jovi, Abigail Breslin, Ludacris, Robert De Niro\ldots'' \textcolor{gray}{\texttt{</knowledge>}} \\[3pt]
& 2
  & \textcolor{teal}{\texttt{<think>}} ``New Year's Eve'' is a romantic comedy starring Ludacris. Now I need to find if any cast member was from ``7th Heaven.'' \textcolor{teal}{\texttt{</think>}} \\
& & \textcolor{blue}{\texttt{<query>}} Who starred in New Year's Eve besides Ludacris? \textcolor{blue}{\texttt{</query>}} \\
& & \textcolor{gray}{\texttt{<knowledge>}} (same New Year's Eve cast list) \textcolor{gray}{\texttt{</knowledge>}} \\[3pt]
& 3
  & \textcolor{teal}{\texttt{<think>}} Halle Berry starred in ``New Year's Eve'' alongside Ludacris. \underline{Halle Berry was a star of ``7th Heaven.''} The director is Garry Marshall. \textcolor{teal}{\texttt{</think>}} \\
& & \textcolor{red!70!black}{\texttt{<answer>}} Garry Marshall \textcolor{red!70!black}{\texttt{</answer>}} \\[2pt]
& & \colorbox{gray!10}{\scriptsize \textbf{F1\,=\,1.000} ~|~ \textbf{PO\,=\,0.072} ~|~ \textbf{Turns\,=\,3} ~|~ Correct answer via \textbf{hallucinated bridge} (Halle Berry $\neq$ 7th Heaven)} \\
\midrule
Ours & 1
  & \textcolor{teal}{\texttt{<think>}} To find the director, I first need to identify which actor from ``7th Heaven'' appeared in a romantic comedy with Ludacris. \textcolor{teal}{\texttt{</think>}} \\
& & \textcolor{blue}{\texttt{<query>}} Which actors from ``7th Heaven'' starred in a romantic comedy with Ludacris? \textcolor{blue}{\texttt{</query>}} \\
& & \textcolor{gray}{\texttt{<knowledge>}} ``She is best known for roles as Sandy Jameson in 7th Heaven\ldots'' / ``Barret Swatek played Cheryl on the WB series 7th Heaven for three seasons\ldots'' \textcolor{gray}{\texttt{</knowledge>}} \\[3pt]
& 2
  & \textcolor{teal}{\texttt{<think>}} Barret Swatek starred in ``7th Heaven.'' She might be the connection. \textcolor{teal}{\texttt{</think>}} \\
& & \textcolor{blue}{\texttt{<query>}} Who directed the romantic comedy featuring Barret Swatek and Ludacris? \textcolor{blue}{\texttt{</query>}} \\
& & \textcolor{gray}{\texttt{<knowledge>}} ``Barret Swatek\ldots is an American actress and comedian who has appeared in Lethal Weapon 4, The 40-Year-Old Virgin\ldots'' / ``New Year's Eve stars Halle Berry, Jessica Biel\ldots Ludacris\ldots'' \textcolor{gray}{\texttt{</knowledge>}} \\[3pt]
& 3
  & \textcolor{teal}{\texttt{<think>}} Barret Swatek starred alongside Ludacris. The director is Garry Marshall. \textcolor{teal}{\texttt{</think>}} \\
& & \textcolor{red!70!black}{\texttt{<answer>}} Garry Marshall \textcolor{red!70!black}{\texttt{</answer>}} \\[2pt]
& & \colorbox{gray!10}{\scriptsize \textbf{F1\,=\,1.000} ~|~ \textbf{PO\,=\,0.111} ~|~ \textbf{Turns\,=\,3} ~|~ Correct reasoning chain; \textbf{valid bridge entity from evidence}} \\
\bottomrule
\end{tabular}
\end{table*}

%% file: case_study/case_007_governor_salary.tex

\begin{table*}[t]
\centering
\caption{Case study on a 2-hop question from MuSiQue. \textbf{Question:} ``What is the salary of the governor of the state Kevin Sessums was born in?'' \textbf{Ground truth:} \$122,160. GRPO produces the exact answer but the salary figure never appears in any retrieved passage (PO\,=\,0.028), indicating parametric hallucination. Ours retrieves the salary from evidence in 3 turns (PO\,=\,0.139).}
\label{tab:case_salary}
\fontsize{7.5pt}{9pt}\selectfont
\setlength{\tabcolsep}{3pt}
\begin{tabular}{p{1.2cm} p{0.4cm} p{12cm}}
\toprule
\textbf{Method} & \textbf{Turn} & \textbf{Content} \\
\midrule
Base & 1
  & \textcolor{teal}{\texttt{<think>}} First, I need to find out which state Kevin Sessums was born in. \textcolor{teal}{\texttt{</think>}} \\
& & \textcolor{blue}{\texttt{<query>}} What state was Kevin Sessums born in? \textcolor{blue}{\texttt{</query>}} \\
& & \textcolor{gray}{\texttt{<knowledge>}} ``Tennessee is a southeastern U.S.\ state\ldots the Mississippi River forms the western border\ldots'' \textcolor{gray}{\texttt{</knowledge>}} \\[3pt]
& 2
  & \textcolor{teal}{\texttt{<think>}} Kevin Sessums was born in Forest, Mississippi. Now I need to find the current governor of Mississippi and their salary. \textcolor{teal}{\texttt{</think>}} \\
& & \textcolor{blue}{\texttt{<query>}} Who is the current governor of Mississippi and what is their salary? \textcolor{blue}{\texttt{</query>}} \\
& & \textcolor{gray}{\texttt{<knowledge>}} ``The 2019 Mississippi gubernatorial election\ldots Phil Bryant barred from seeking a third term\ldots'' \textcolor{gray}{\texttt{</knowledge>}} \\[3pt]
& \multicolumn{2}{l}{\quad\textcolor{gray!60}{\dotfill~\textit{Turns 3--6: repeated ``governor of Mississippi salary'' queries; no salary figure returned}~\dotfill}} \\[3pt]
& 7
  & \textcolor{teal}{\texttt{<think>}} The information does not provide the current governor or their salary. \textcolor{teal}{\texttt{</think>}} \\
& & \textcolor{red!70!black}{\texttt{<answer>}} Not available \textcolor{red!70!black}{\texttt{</answer>}} \\[2pt]
& & \colorbox{gray!10}{\scriptsize \textbf{F1\,=\,0.000} ~|~ \textbf{PO\,=\,0.030} ~|~ \textbf{Turns\,=\,7} ~|~ Stuck in retrieval loop; gives up} \\
\midrule
GRPO & 1
  & \textcolor{teal}{\texttt{<think>}} I need to first find out which state Kevin Sessums was born in. \textcolor{teal}{\texttt{</think>}} \\
& & \textcolor{blue}{\texttt{<query>}} What state was Kevin Sessums born in? \textcolor{blue}{\texttt{</query>}} \\
& & \textcolor{gray}{\texttt{<knowledge>}} ``Tennessee is a southeastern U.S.\ state\ldots'' \textcolor{gray}{\texttt{</knowledge>}} \\[3pt]
& 2
  & \textcolor{teal}{\texttt{<think>}} Kevin Sessums was born in Forest, Mississippi. Now I need to find the current governor of Mississippi and their salary. \textcolor{teal}{\texttt{</think>}} \\
& & \textcolor{blue}{\texttt{<query>}} Who is the current governor of Mississippi and what is their salary? \textcolor{blue}{\texttt{</query>}} \\
& & \textcolor{gray}{\texttt{<knowledge>}} ``The 2019 Mississippi gubernatorial election\ldots Phil Bryant barred from seeking a third term\ldots'' \textcolor{gray}{\texttt{</knowledge>}} \\[3pt]
& \multicolumn{2}{l}{\quad\textcolor{gray!60}{\dotfill~\textit{Turns 3--7: repeated salary queries; no salary figure returned}~\dotfill}} \\[3pt]
& 8
  & \textcolor{teal}{\texttt{<think>}} I have searched multiple times but could not find the salary. \textcolor{teal}{\texttt{</think>}} \\
& & \textcolor{red!70!black}{\texttt{<answer>}} \$122,160 \textcolor{red!70!black}{\texttt{</answer>}} \\[2pt]
& & \colorbox{gray!10}{\scriptsize \textbf{F1\,=\,1.000} ~|~ \textbf{PO\,=\,0.028} ~|~ \textbf{Turns\,=\,8} ~|~ Correct answer from \textbf{parametric memory; no evidence retrieved}} \\
\midrule
Ours & 1
  & \textcolor{teal}{\texttt{<think>}} I need to determine which state Kevin Sessums was born in. \textcolor{teal}{\texttt{</think>}} \\
& & \textcolor{blue}{\texttt{<query>}} Kevin Sessums birthplace \textcolor{blue}{\texttt{</query>}} \\
& & \textcolor{gray}{\texttt{<knowledge>}} ``Kevin Sessums, born in 1956 in Forest, \textbf{Mississippi}, is the brother of artist Dr.\ J.\ Kim Sessums of Brookhaven, Mississippi.'' \textcolor{gray}{\texttt{</knowledge>}} \\[3pt]
& 2
  & \textcolor{teal}{\texttt{<think>}} Kevin Sessums was born in Forest, Mississippi. Now I need to find the governor's salary. \textcolor{teal}{\texttt{</think>}} \\
& & \textcolor{blue}{\texttt{<query>}} current governor of Mississippi and salary \textcolor{blue}{\texttt{</query>}} \\
& & \textcolor{gray}{\texttt{<knowledge>}} ``Phil Bryant has been the incumbent\ldots annual salary in 2013 was \textbf{\$122,160}.'' \textcolor{gray}{\texttt{</knowledge>}} \\[3pt]
& 3
  & \textcolor{teal}{\texttt{<think>}} The current governor of Mississippi is Phil Bryant, and his annual salary in 2013 was \$122,160. \textcolor{teal}{\texttt{</think>}} \\
& & \textcolor{red!70!black}{\texttt{<answer>}} \$122,160 \textcolor{red!70!black}{\texttt{</answer>}} \\[2pt]
& & \colorbox{gray!10}{\scriptsize \textbf{F1\,=\,1.000} ~|~ \textbf{PO\,=\,0.139} ~|~ \textbf{Turns\,=\,3} ~|~ Correct answer \textbf{grounded in retrieved evidence}} \\
\bottomrule
\end{tabular}
\end{table*}